\documentclass[11pt]{article}

\usepackage[preprint]{acl}

\usepackage{times}
\usepackage{tabularx}
\usepackage{latexsym}
\usepackage{amsmath}
\usepackage{booktabs}
\usepackage{multirow}
\usepackage{float}
\usepackage{adjustbox}
\usepackage{placeins} 
\usepackage[most]{tcolorbox}
\usepackage{caption}
\RequirePackage{tikz}
\usetikzlibrary{shapes, backgrounds}
\tcbuselibrary{skins,breakable,listings}
\usepackage{enumitem}
\usepackage{caption}
\usepackage{tcolorbox}
\usepackage[T1]{fontenc}
\usepackage[utf8]{inputenc}

\usepackage{microtype}

\usepackage{inconsolata}

\usepackage{graphicx}

\title{ReGraM: Region-First Knowledge Graph Reasoning for Medical Question Answering}

\author{
\textbf{Chaerin Lee}$^{*}$ \quad
Sohee Park$^{\dagger}$ \quad
Hyunsik Na$^{\dagger}$  \quad
Daseon Choi$^{\ddagger}$ \\
Department of Software, Soongsil University, Seoul, Republic of Korea \\
\ttfamily
\{chaerin1112, sosohi, rnrud7932\}@soongsil.ac.kr \quad sunchoi@ssu.ac.kr
}

\usepackage{needspace}
\begin{document}
\sloppy 
\maketitle
%==================================================abstract========================================
\begin{abstract}
Recent studies in medical question answering (Medical QA) have actively explored the integration of large language models (LLMs) with biomedical knowledge graphs (KGs) to improve factual accuracy.
However, most existing approaches still rely on traversing the entire KG or performing large-scale retrieval, which introduces substantial noise and leads to unstable multi-hop reasoning.
We argue that the core challenge lies not in expanding access to knowledge, but in identifying and reasoning over the appropriate subset of evidence for each query. \textbf{ReGraM} is a region-first knowledge graph reasoning framework that addresses this challenge by constructing a query-aligned subgraph and performing stepwise reasoning constrained to this localized region under multiple evidence-aware modes.
By focusing inference on only the most relevant portion of the KG, ReGraM departs from the assumption that all relations are equally useful—an assumption that rarely holds in domain-specific medical settings.
Experiments on seven medical QA benchmarks demonstrate that ReGraM consistently outperforms a strong baseline (KGARevion), achieving an 8.04\% absolute accuracy gain on MCQ, a 4.50\% gain on SAQ, and a 42.9\% reduction in hallucination rate.
Ablation and qualitative analyses further show that aligning region construction with hop-wise reasoning is the primary driver of these improvements.
Overall, our results highlight region-first KG reasoning as an effective paradigm for improving factual accuracy and consistency in medical QA.
\end{abstract}

% =========================================================1장!!! Introduction==================================================================
\section{Introduction}

Medical question answering (Medical QA) increasingly requires multi-hop reasoning that connects symptoms, mechanisms, diagnoses, treatments, and molecular interactions \citep{xie2024preliminary, singhal2025toward, wu2025medical}.

Despite this growing demand, most existing LLM-based Medical QA systems still rely on shallow, retrieval-based reasoning. Even methods designed for multi-hop inference often lack explicit structural constraints, causing reasoning chains to drift across unrelated biomedical concepts.
As a result, clinical evaluations frequently report medical hallucinations—plausible but unsupported statements that pose serious safety risks in real-world applications \citep{saab2024capabilities, williams2024evaluating}.
To address these issues, recent work has integrated large language models with biomedical KGs through KG-based QA, KG-RAG, and graph-constrained prompting \citep{zeng2025kosel, sui2025fidelis, shi2025mkrag, long2024bailicai}.
Methods such as KGARevion further apply post-hoc verification of LLM-generated triplets against a KG \citep{jeong2024improving}.
However, most KG-based approaches still treat the KG as a flat search space spanning heterogeneous biomedical subdomains and relation types \citep{Chandak2023KGPrecisionMedicine}.
Because typical queries require only a small, task-relevant subset of the KG, global traversal introduces substantial retrieval noise that accumulates over multi-hop reasoning and destabilizes inference \citep{su2024kgarevion, jeong2024improving, wu2025medical}.

We propose a different design principle: rather than exploring the entire KG prior to reasoning, we first construct a semantically aligned subgraph tailored to the query and constrain all subsequent reasoning within that region.
Unlike post-hoc verification or soft context filtering, this region is treated as a hard structural boundary throughout multi-hop inference,
preventing semantically irrelevant relations from intervening once reasoning begins.
We instantiate this principle in a framework called \textbf{ReGraM}, which performs query typing, region selection, evidence-aware reasoning, and verification entirely within a localized KG region.
This design enforces early restriction of the reasoning space and promotes factual consistency by structurally excluding irrelevant relations.

We evaluate ReGraM on PrimeKG using the same entity set and relation schema as KGARevion to ensure a controlled comparison.
To improve region construction, we refine the relation description map used for retrieval and prompting, without modifying the underlying KG.

Although evaluated in the medical domain, the region-first principle generalizes to other structured domains with high relational complexity. Our main contributions are as follows:
\begin{enumerate}[leftmargin=*, itemsep=0pt, topsep=0pt, parsep=0pt, partopsep=0pt]
    \item We propose \textbf{ReGraM}, a region-first KG reasoning framework that constructs a query-aligned subgraph and performs reasoning via multi-hop decomposition within it. This design structurally constrains the reasoning space and mitigates semantic drift by excluding irrelevant relations.
    \item We demonstrate substantial empirical gains: an 8.04\% improvement in MCQ accuracy, a 4.50\% improvement in SAQ accuracy, a 32--48\% reduction in hallucination rates, and a 46\% reduction in inference latency across seven medical QA benchmarks and multiple hallucination testbeds.
    \item We conduct ablation and error analyses showing that region selection and multi-hop decomposition are critical to factual consistency, and we validate the contribution of each module to overall system stability.
\end{enumerate}

% =================================================================2장!!! - Related Work==================================================
\section{Related Work}
% ===============================================2.1절!!!! - LLM-based Reasoning and Multi-hop Inference===========================================
\subsection{LLM-based Reasoning and Multi-hop Inference}

Medical question answering (MedQA) requires multi-hop reasoning that connects diverse biomedical concepts in a stepwise manner, which is challenging for single-step retrieval-based approaches.
To address this challenge, a wide range of LLM-based reasoning methods have been proposed, including Chain-of-Thought and its variants, ReAct, Reflexion, SelfCheck, and STaR \citep{wei2022chain, yao2022react, shinn2023reflexion, manakul2023selfcheckgpt, zelikman2024star}.
However, these methods typically rely on the LLM’s internal knowledge without external structural constraints, often leading to reasoning drift, cumulative errors, and factual hallucinations in complex biomedical scenarios \citep{bang-etal-2023-multitask, 10.1145/3703155}.

In particular, many MedQA tasks require connecting multiple entities across relational chains.
LLMs alone struggle to maintain consistency across these hops, highlighting the need to incorporate external structured knowledge—such as knowledge graphs (KGs)—to provide relational grounding and guide multi-hop inference \citep{chen2019multi, luo2023reasoning, sakarvadia2024towards}.

%===========================================그림 1!!!! - architecture ===========================================
\begin{figure*}[!t]
\centering
\includegraphics[width=\textwidth]{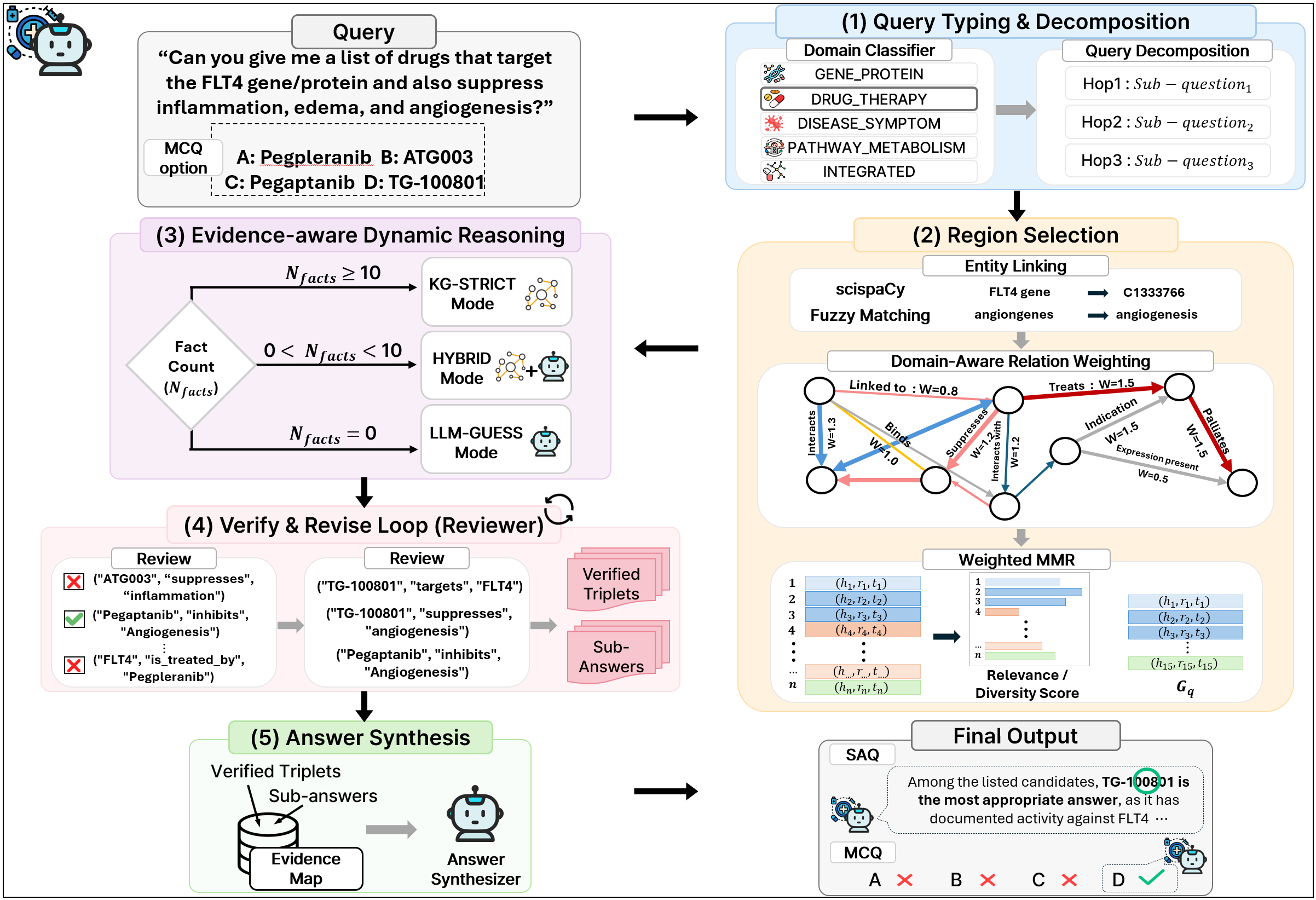}
\caption{Overview of the ReGraM framework.
Given a medical query, ReGraM constructs a query-aligned local knowledge graph region and performs reasoning exclusively within the selected region to generate the final answer.}
\label{fig:architecture}
\end{figure*}

% ===============================================2.2절!!!! - KG-based Reasoning for Medical QA ===========================================
\subsection{KG-based Reasoning for Medical QA}

To improve factual grounding, a variety of models have explored knowledge graph-based QA in the biomedical domain, including QAGNN \citep{yasunaga2021qa}, JointLK \citep{sun-etal-2022-jointlk}, KagNet \citep{lin-etal-2019-kagnet}, MedReason \citep{wu2025medical}, and ontology-guided KGQA \citep{liu-etal-2025-ontology}.
While effective in incorporating KG structure, these methods typically operate over the entire graph without explicitly tailoring the reasoning space to the query context.

In large-scale biomedical KGs such as PrimeKG, full-graph traversal introduces irrelevant relations, propagates retrieval noise, and fragments evidence due to sparsity \citep{zhai2024towards, mavi-etal-2023-retrieval, Chandak2023KGPrecisionMedicine}.
Some recent methods address this issue by incorporating verification mechanisms (e.g., KGARevion \citep{su2024kgarevion}) or by organizing retrieval contexts using graph structure (e.g., GraphRAG \citep{xiao2025graphrag} and MedGraphRAG \citep{wu2025medical}).
However, these approaches often treat KG retrieval as a preprocessing step, leaving the subsequent reasoning process unconstrained beyond the retrieved context.
As a result, retrieval noise and semantically irrelevant relations can still influence downstream reasoning, especially in multi-hop and generative settings.
This issue is particularly pronounced in generative QA, where early retrieval errors can propagate and compound across reasoning steps.

In contrast, our work imposes an explicit structural boundary \emph{before} reasoning begins.
Rather than applying KG pruning post hoc or relying on soft context filtering, we define a query-specific subgraph as the primary reasoning space, ensuring that multi-hop inference remains structurally grounded throughout the entire process.

% ==================================================================== 3장!!! - Method ===============================================================
\section{Method}

Figure~\ref{fig:architecture} presents an overview of ReGraM.
ReGraM is a region-first knowledge graph reasoning framework that constructs a query-aligned subgraph and performs multi-hop reasoning under a hard region constraint.

This section describes the overall framework and its core components.
Additional implementation details, prompt templates, and qualitative examples are provided in the appendix.

% =============================================3.1절!!! - Query Typing and CoT-based Decomposition===============================================
\subsection{Query Typing and Decomposition}

ReGraM begins by analyzing the structure of the input query to guide subsequent region construction.
Given an input query $Q$, the model first classifies it into one of five high-level medical domains
(\texttt{GENE/PROTEIN}, \texttt{DRUG/THERAPY}, \texttt{DISEASE/SYMPTOM}, \texttt{PATHWAY/METABOLISM}, or \texttt{INTEGRATED})
using an LLM-based classification prompt.
The predicted domain serves as a semantic prior for adjusting relation importance during KG region selection.

The query is then decomposed into up to three sub-questions using a one-shot CoT prompt,
aligning the reasoning process with the underlying multi-hop structure
(Appendix~\ref{app:appendix_A} and Figure~\ref{fig:prompt2}).

% ===============================================================3.2절!!! - Region Selection =======================================================
\subsection{Region Selection}

For each sub-question $q_i$, ReGraM constructs a localized KG region rather than retrieving from the full graph.
Medical entities are extracted using the UMLS linker in SciSpaCy \citep{neumann-etal-2019-scispacy, bodenreider2004unified}
and expanded into an enriched entity set $E_{\text{exp}}$
via UMLS synonyms, custom alias dictionaries curated from PrimeKG node attributes,
and fuzzy string matching (similarity $\geq 90$) with RapidFuzz \citep{ye2021}.

Candidate triplets $T_{\text{cand}}$ connected to $E_{\text{exp}}$ are collected from the KG and reweighted using domain-specific relation weights
$w_r \in \{0.5, 0.8, 1.0, 1.2, 1.5\}$, which encode the relevance of each relation type to the predicted query domain.
To prioritize both semantic relevance and evidence diversity, triplets are re-ranked using a weighted
Maximal Marginal Relevance (MMR) criterion \citep{10.1145/290941.291025}:

\begin{equation}
\label{eq:mmr}
\begin{aligned}
\text{MMR}(t)
&= \lambda \, \text{Sim}(q_i, t) \, w_r \\
&\quad - (1-\lambda)\, \max_{t' \in S} \text{Sim}(t, t')
\end{aligned}
\end{equation}

Here, $\text{Sim}(\cdot,\cdot)$ denotes the cosine similarity between Sentence-BERT embeddings
(\texttt{all-MiniLM-L6-v2}) of the sub-question $q_i$ and the textual representation of a triplet $t$ \citep{reimers2019sentence}.
The second term penalizes redundancy by discouraging selection of triplets that are semantically similar to previously selected ones,
where $S$ denotes the set of triplets already chosen for the current region.
The hyperparameter $\lambda = 0.7$ balances relevance and diversity during region construction.

The top-$K$ ranked triplets (with $K=15$) constitute the query-specific KG region
$G_q = (V_q, E_q)$, where $E_q$ is the selected triplet set and
$V_q$ contains all head and tail entities appearing in $E_q$.
All subsequent knowledge access operations during reasoning are restricted to this region,
formally denoted as $\text{Lookup}(q_i) \subseteq G_q$.
This restriction enforces region-constrained reasoning by preventing the model from accessing relations outside the query-aligned subgraph
(see Appendix~\ref{app:appendix_B}).

ReGraM does not modify the underlying KG; performance gains stem from structurally bounding inference within a query-aligned subgraph.
This design enables the model to more effectively interpret and utilize the semantic structure already encoded in PrimeKG,
for example by distinguishing clinically relevant relations from generic associations during region construction.
% ======================================================3.3절!!! - Evidence-aware Dynamic Reasoning =================================================
\subsection{Evidence-aware Dynamic Reasoning}

ReGraM adapts its reasoning behavior based on the amount of retrieved evidence.
We unify the notion of hop count across modules by fixing the maximum number of reasoning steps to $H = 3$,
implemented as up to three decomposed sub-questions.
No additional graph-walk or in-region traversal depth is introduced beyond these steps;
all reasoning is organized within this fixed hop budget.

Let $N_{\text{facts}}$ denote the number of region triplets available for a given sub-question.
Based on $N_{\text{facts}}$, the model dynamically switches among three reasoning modes.
Importantly, even when hypothesis generation is permitted, reasoning remains constrained.
Across all modes, ReGraM operates under a closed-world assumption:
entities must belong to $V_q$, and relations must be selected from a predefined schema set $\mathcal{R}$.

\textbf{KG-STRICT} relies solely on region evidence when sufficient facts are available,
while \textbf{LLM-GUESS} falls back to parametric knowledge when no region evidence exists.
\textbf{HYBRID} performs constrained hypothesis completion using only in-region entities and schema-allowed relations,
followed by Reviewer verification, enabling limited generalization when KG evidence is partial but non-empty. All reasoning steps are executed strictly within the selected region $G_q$,
with the total number of reasoning steps fixed to $H = 3$ for stability.

%원래 표1자리
% ======================================================3.4절!!! - Verify-and-Revise Loop======================================================
\subsection{Verify \& Revise Loop}

In \textbf{HYBRID} mode, ReGraM employs a verification mechanism to control hallucinated or inconsistent hypotheses.
Generated triplets are evaluated by a LoRA-tuned Llama-3.1-8B binary classifier, referred to as the \textbf{Reviewer} \citep{hu2022lora, grattafiori2024llama}.
The Reviewer assigns a score $s_{\theta}$ to each triplet and accepts it as valid if $s_{\theta} \ge 0.5$.

Triplets rejected by the Reviewer are not discarded immediately; instead, they are passed to a revision prompt that attempts to correct semantic or schema-level inconsistencies.
The revised triplets are then re-evaluated by the Reviewer.
This verification--revision loop runs for up to two iterations, and only Reviewer-approved triplets are incorporated into the reasoning evidence
(see Appendix~\ref{app:appendix_A} and Figure~\ref{fig:prompt4}).

% ======================================================3.5절 !!! - Evidence-grounded Answer Synthesis==============================================
\subsection{Answer Synthesis}

Throughout reasoning, ReGraM stores intermediate sub-answers and supporting evidence in an \textbf{Evidence Map}.
Final answer generation is performed using mode-specific synthesis templates:
\textbf{KG-STRICT} produces concise factual answers grounded solely in region evidence;
\textbf{HYBRID} generates stepwise explanations based on Reviewer-verified triplets;
and \textbf{LLM-GUESS} explicitly indicates missing KG evidence while providing a plausible answer.

The \textbf{Answer Synthesizer} integrates hop-level outputs into a coherent final response grounded in the accumulated evidence
(see Appendix~\ref{app:appendix_A}, Figures~\ref{fig:prompt5}--\ref{fig:prompt8}).
Taken together, ReGraM’s region-first design separates \emph{where} to reason (region construction)
from \emph{how} to reason (evidence-aware inference), allowing reasoning behavior to adapt to evidence availability.

% ======================================================표1 !!! - main_result & 데이터셋별(SAQ만) hallucination 발생률 ======================================================
\begin{table*}[t]
\centering
\footnotesize
\setlength{\tabcolsep}{4.5pt}
\renewcommand{\arraystretch}{1.15}

\begin{tabular}{cllcccccccc}
\toprule
\textbf{Task} & \textbf{Metric} & \textbf{Model} &
\textbf{Basic} & \textbf{Inter.} & \textbf{Expert} &
\textbf{MedQA} & \textbf{Afrimed} & \textbf{MMLU} &
\textbf{PubMed} & \textbf{Avg} \\
\midrule

% ================= MCQ =================
\multirow[c]{3}{*}{MCQ}
& \multirow[c]{3}{*}{ACC $\uparrow$}
& KGARevion
& 38.8 & 35.0 & 33.8 & 51.7 & 49.9 & 62.8 & 54.8 & 46.2 \\
&& \textbf{ReGraM}
& \textbf{47.1} & \textbf{45.0} & \textbf{45.9}
& \textbf{54.4} & \textbf{56.5} & \textbf{65.7}
& \textbf{64.9} & \textbf{54.2} \\
&& \multicolumn{1}{c}{$\Delta$}
& +8.3 & +9.9 & +12.1 & +2.7 & +6.6 & +2.9 & +10.1 & +8.0 \\

\midrule

% ================= SAQ (ACC + IS together) =================
\multirow[c]{6}{*}{SAQ}
& \multirow[c]{3}{*}{ACC $\uparrow$}
& KGARevion
& 10.2 & 13.9 & 14.8 & 36.5 & 26.1 & 65.7 & 53.5 & 31.5 \\
&& \textbf{ReGraM}
& \textbf{11.3} & \textbf{13.9} & \textbf{20.8}
& \textbf{51.0} & \textbf{47.1} & 65.5
& 44.7 & \textbf{36.0} \\
&& \multicolumn{1}{c}{$\Delta$}
& +1.1 & 0.0 & +6.0 & +14.5 & +21.0 & -0.3 & -8.8 & +4.5 \\

\cmidrule(lr){2-11}

&\multirow[c]{3}{*}{IS $\downarrow$}
& KGARevion
& 57.5 & 53.8 & 57.6 & 52.9 & 23.7 & 41.8 & 36.0 & 46.2 \\
&& \textbf{ReGraM}
& \textbf{30.5} & \textbf{32.0} & \textbf{30.1}
& \textbf{28.2} & \textbf{16.1} & \textbf{23.8}
& \textbf{24.4} & \textbf{26.4} \\
&& \multicolumn{1}{c}{$\Delta_{\text{IS}}$}
& 27.0 & 40.5 & 47.7 & 46.7 & 32.1 & 43.1 & 32.2 & 42.9 \\

\bottomrule
\end{tabular}

\caption{
Performance of \textbf{ReGraM} and KGARevion on seven medical QA benchmarks.
ACC denotes accuracy (\%), and IS denotes the hallucination inconsistency score evaluated only for SAQ (lower is better).
$\Delta$ indicates absolute accuracy improvement (percentage points), and $\Delta_{\text{IS}}$ denotes relative reduction for IS (\%),
computed as $(\text{IS}_{\text{KGARevion}} - \text{IS}_{\text{ReGraM}}) / \text{IS}_{\text{KGARevion}} \times 100$.
}
\label{tab:main_result}
\end{table*}

% ======================================================4장!!! -  Experiments ======================================================
\section{Experiments}

This section analyzes how region-first KG reasoning affects medical QA accuracy, hallucination robustness, and reasoning efficiency. 
All experiments compare ReGraM with the baseline KGARevion under identical settings.
Both methods operate on the same biomedical knowledge graph, PrimeKG, with the same entity set and relation schema.

% ======================================================4.1절!! Experimental Setup!!!======================================================
\subsection{Experimental Setup}
\label{subsec:exp_setup}

Both ReGraM and KGARevion are based on \texttt{Llama-3.1-8B} and operate on the same
biomedical knowledge graph, PrimeKG, under identical training and inference settings.
No additional knowledge graphs, entities, relations, or edges are introduced in ReGraM.

The key difference lies in how relational knowledge is utilized.
ReGraM extends the relation description map originally used in KGARevion
to more comprehensively cover the diverse relation types present in PrimeKG.
This extension does not alter the knowledge available to the model,
but ensures that domain-relevant relations are not systematically underweighted
when the reasoning space is explicitly constrained.
As a result, region construction more accurately aligns retrieved triplets
with the query intent, while preserving a fully controlled KG setting.
Performance gains in ReGraM are not attributable to increased knowledge coverage,
as the underlying knowledge graph remains unchanged.
Instead, improvements stem from semantically interpreting and structurally constraining
existing PrimeKG relations during region construction and reasoning.

Traditional KG-QA models such as QAGNN, JointLK, and KagNet assume supervised QA training
and differ fundamentally from our generative KG-alignment objective.
ReGraM is positioned as a structural extension of KGARevion under identical training and inference settings;
accordingly, we focus on a controlled comparison to isolate the effect of region-first bounding.
Broader comparisons to KG-RAG or GraphRAG-style baselines are left to future work.
Experiments were conducted on seven medical QA benchmarks: MedDDx-Basic, MedDDx-Intermediate,
MedDDx-Expert, MedQA, PubMedQA, MMLU-Medical, and AfrimedQA.
For readability, we refer to the three MedDDx variants as Basic, Intermediate, and Expert
in tables and figures.
Overall, region-first reasoning improves stability under the semantic diversity of PrimeKG relations.

% ======================================================4.2절!!! Evaluation Protocols!!!======================================================
\subsection{Evaluation Protocols}

We evaluate model performance along two axes: \textbf{accuracy} and \textbf{hallucination robustness}.
Standard medical QA benchmarks are assessed under both multiple-choice (MCQ) and short-answer (SAQ) settings,
while hallucination robustness is evaluated using the MedHallu and Med-HALT benchmarks.

\paragraph{Standard Medical QA Benchmarks (MCQ \& SAQ)}
In the MCQ setting, each model receives a question with candidate options and outputs an answer choice,
and we report exact-match accuracy.
In the SAQ setting, the model generates an open-ended answer.
Because embedding similarity alone does not reliably capture biomedical correctness,
we employ an LLM-as-a-Judge protocol \citep{badshah2025reference, chang2024survey} using GPT-4o
to assess semantic alignment between generated and reference answers.
Following common practice, outputs with a similarity score of at least 0.8 are treated as correct,
and the same judge and threshold are used for both methods to ensure a controlled comparison.
Evaluation prompts are provided in Appendix~\ref{app:appendix_A} and Figure~\ref{fig:prompt9}.

% ======================================================Hallucination Benchmarks (MedHallu, Med-HALT)=================================================

\paragraph{Hallucination Benchmarks (MedHallu, Med-HALT)}
MedHallu \citep{10.1145/3703155} measures factual accuracy, hallucination rate, and reasoning error rate in free-form generation.  
Med-HALT \citep{wu2025medical}, on the other hand, evaluates model stability under adversarial conditions through three sub-tests: Fake Question Test (FQT), None-of-the-Above (NOTA), and False Confidence Test (FCT).  

Our LLM-based inconsistency analysis and MedHallu evaluation follow the SAQ (open-ended) setting, while Med-HALT adopts the MCQ (multiple-choice) discriminative format.
Evaluation prompts and parsing rules are provided in Appendix~\ref{app:appendix_A} and Figure~\ref{fig:prompt11}.

For MedHallu’s semantic alignment score (MH-Sim), we use GPT-4o to measure the similarity between the generated answer and the ground-truth answer on a 0--100 scale, and report the average score across samples.

%==================== 표 2!!! Domain-wise IS vs KG Coverage (Main) ====================
\begin{table}[t]
\centering
\small
\begin{tabular}{lcc}
\toprule
\textbf{Domain} & \textbf{KG Coverage (\%)} & \textbf{Mean IS} \\
\midrule
Drug\_Therapy   & 68.14 & 5.08 \\
Gene\_Protein   & 60.30 & 5.97 \\
Disease\_Symptom& 15.66 & 7.16 \\
Pathway\_Metab. & 26.31 & 5.42 \\
Integrated      & 83.52 & 6.83 \\
\bottomrule
\end{tabular}
\caption{Domain-wise hallucination inconsistency score (IS) vs KG coverage. Higher coverage generally leads to lower IS.}
\label{tab:domain_is}
\end{table}

%====================================================== 4.3절!!! Main Results on Medical QA Performance=======================================
\subsection{Main Results on Medical QA Performance}

Table~\ref{tab:main_result} summarizes the comparative performance of ReGraM and KGARevion
across seven medical QA benchmarks under both multiple-choice (MCQ) and short-answer (SAQ) settings.
We report absolute accuracy improvements ($\Delta$, percentage points) as well as relative reductions
in inconsistency score ($\Delta_{\text{IS}}$, percentage).
Overall, ReGraM achieves consistent gains over the baseline, improving MCQ accuracy by 8.04\%
and SAQ accuracy by 4.50\%, with particularly strong improvements on datasets requiring intensive multi-hop reasoning.

ReGraM shows the largest gains on the MedDDx benchmarks (Basic, Intermediate, and Expert),
all of which involve compositional reasoning across multiple biomedical entities.

In the SAQ setting, ReGraM yields substantial improvements on MedQA (14.50\%)
and AfrimedQA (21.04\%), both of which involve complex clinical scenarios that demand precise evidence alignment.
These results indicate that integrating semantic region selection with dynamic reasoning modes
effectively improves open-ended answer generation.

A performance trade-off is observed on PubMedQA (-8.85\%), where KG evidence is sparse
or weakly aligned with question semantics.
This behavior reflects a design trade-off rather than a failure case:
in datasets with limited KG coverage, aggressive region restriction and conservative verification
can under-retrieve relevant evidence needed for answer generation.

Results in Table~\ref{tab:main_result} are averaged over three runs (mean $\pm$ std).
Observed improvements consistently exceed baseline variance,
supporting the effectiveness of the region-first design
in enhancing accuracy and compositional reasoning across diverse medical QA tasks.

% ======================================================4.4절!!! Hallucination Evaluation Results==============================================
\subsection{Hallucination Evaluation Results}

We evaluate hallucination robustness using IS (SAQ), MedHallu (generative), and Med-HALT (adversarial MCQ).

%============Dataset-wise Hallucination Reduction(GPT-5 IS) dataset_ir, dataset_ir_figure ==================================
\paragraph{Dataset-wise Hallucination Reduction (IS)}
Table~\ref{tab:main_result} shows factual hallucination inconsistency scores (IS) across datasets under the SAQ setting, scored using an LLM-based inconsistency evaluator (see Appendix~\ref{app:appendix_A}).  
ReGraM reduces the average IS from 46.2 to 26.4, achieving a 42.9\% reduction overall. 
The largest reductions are observed in multi-hop–intensive datasets such as MedDDx-Basic (–47.0\%), MedDDx-Intermediate (–40.5\%), and MedDDx-Expert (–47.8\%), demonstrating that ReGraM is particularly effective where multi-step reasoning is required.  
This indicates that region bounding reduces drift, and the Reviewer suppresses unsupported steps.

%====================================================== Domain-wise Hallucination Robustness and KG !!!=======================================
\paragraph{Domain-wise Hallucination Robustness}
Table~\ref{tab:domain_is} reports mean hallucination inconsistency scores (IS)
across five biomedical domains under the SAQ setting.
Overall, higher KG coverage tends to correspond to lower hallucination rates;
for instance, the Drug\_Therapy domain exhibits the lowest IS.
However, this relationship is not uniform across domains.
Notably, the \texttt{Integrated} domain shows relatively high IS despite having the highest KG coverage.

We further conduct a finer-grained analysis in Appendix~\ref{app:domain-wise_hallucination},
demonstrating that hallucination robustness also depends on additional structural factors,
including relation granularity, prompt design, and evidence redundancy.

% ======================================================표3!!! - medhallu===================================================================
\begin{table}[t]
\centering
\small
\begin{tabular}{lcc}
\toprule
\textbf{Metric} & \textbf{KGARevion} & \textbf{ReGraM} \\
\midrule
MH-ACC & 10.00 & \textbf{43.50} \\
MH-Sim & 0.00 & \textbf{54.70} \\
MH-Hallu & 90.00 & \textbf{89.00} \\
MH-RError & 100.00 & \textbf{72.00} \\
\bottomrule
\end{tabular}
\caption{MedHallu results under SAQ. MH-ACC ($\uparrow$): accuracy on a 0--100 scale. MH-Sim ($\uparrow$): semantic similarity to the ground-truth answer on a 0--100 scale. MH-Hallu ($\downarrow$) and MH-RError ($\downarrow$): hallucination and reasoning error rates (\%).}
\label{tab:medhallu}
\end{table}
%====================================================== 표5!!! ablation study 표!!! - ablation ======================================================
\begin{table*}[!t]
\centering
\small
\resizebox{\textwidth}{!}{%
\begin{tabular}{llc|cc|cc}
\toprule
\textbf{Dataset} & \textbf{Type} & \textbf{ReGraM} & \textbf{w/o Domain Prior} & \textbf{$\Delta$ (prior)} & \textbf{w/o Multihop} & \textbf{$\Delta$ (multihop)} \\
\midrule
MedDDx-Expert & MCQ & 45.88 & 39.34 & $+$6.54 & 41.41 & $+$3.03 \\
MedQA & MCQ & 54.38 & 56.48 & $-$2.10 & 55.38 & $-$1.00 \\
MMLU-Med & MCQ & 65.72 & 67.77 & $-$2.05 & 67.95 & $-$2.23 \\
PubMedQA & MCQ & 64.86 & 67.30 & $-$2.44 & 67.30 & $-$5.01 \\
\midrule
MedDDx-Expert & SAQ & 20.81 & 17.18 & $+$3.63 & 18.43 & $+$2.38 \\
MedQA & SAQ & 51.03 & 2.91 & $+$48.12 & 19.80 & $+$29.28 \\
MMLU-Med & SAQ & 65.45 & 5.33 & $+$60.12 & 31.86 & $+$33.59 \\
PubMedQA & SAQ & 44.65 & 41.90 & $+$2.75 & 42.30 & $+$2.35 \\
\bottomrule
\end{tabular}
}
\caption{Ablation results for ReGraM (MCQ and SAQ). Domain prior and multi-hop reasoning are both essential for maintaining factual consistency.}
\label{tab:ablation}
\end{table*}

% ======================================================표4 !!! - medhalt======================================================
\begin{table}[t]
\centering
\small
\begin{tabular}{lcc}
\toprule
\textbf{Metric} & \textbf{KGARevion} & \textbf{ReGraM} \\
\midrule
FQT-ACC & 22.00 & \textbf{24.80} \\
FQT-Hallu & 78.00 & \textbf{76.50} \\
NOTA-ACC & 22.00 & \textbf{19.67} \\
NOTA-Hallu & 78.00 & \textbf{80.33} \\
FCT-ACC & 22.00 & \textbf{25.19} \\
FCT-Hallu & 78.00 & \textbf{74.81} \\
\bottomrule
\end{tabular}
\caption{MedHALT results. ACC ($\uparrow$): accuracy. Hallu ($\downarrow$): hallucination rate.}
\label{tab:medhalt}
\end{table}

% ======================================================MedHallu: Free-form Generative Hallucination !!!=======================================
\paragraph{MedHallu: Free-form Generative Hallucination}
Table~\ref{tab:medhallu} presents results on the MedHallu benchmark.
While KGARevion hallucinates in nearly all generated answers, ReGraM significantly improves factual accuracy, semantic similarity, and reasoning reliability.
These results demonstrate ReGraM’s effectiveness in suppressing hallucinations under free-form generative settings.

% ======================================================Med-HALT: Adversarial Discriminative Hallucination !!!======================================
\paragraph{Med-HALT: Adversarial Discriminative Hallucination}
Table~\ref{tab:medhalt} presents performance on Med-HALT, which probes hallucination under adversarial conditions through Fake-Question (FQT), None-of-the-Above (NOTA), and False-Confidence Test (FCT).
ReGraM outperforms KGARevion in the FCT sub-test ($+$3.19 ACC), indicating stronger factual discrimination under confidence-sensitive evaluation.
However, performance slightly degrades on FQT and NOTA.
This can be attributed to ReGraM’s region-first design: while it discourages unsupported extrapolation, it also biases the model toward selecting the most plausible answer \emph{within} the retrieved region.
As a result, in adversarial settings where the correct response is to abstain (e.g., NOTA), ReGraM may over-commit to weak in-region evidence.
Overall, these results highlight a fundamental trade-off between hallucination suppression and abstention under adversarial inputs.

%==================== 표 6!!!! - Reviewer Ablation Summary (Main) ====================
\begin{table}[t]
\centering
\small
\setlength{\tabcolsep}{5pt}
\begin{tabular}{lcc}
\toprule
\textbf{Metric} & \textbf{ReGraM} & \textbf{w/o Reviewer} \\
\midrule
SAQ Accuracy (avg.) $\uparrow$ & \textbf{36.03} & 33.17 \\
MedHallu Accuracy $\uparrow$ & \textbf{43.50} & 0.00 \\
MedHallu Similarity $\uparrow$ & \textbf{54.70} & 0.00 \\
MedHallu Hallucination $\downarrow$ & \textbf{89.0} & 100.0 \\
MedHallu Reasoning Error $\downarrow$ & \textbf{72.0} & 100.0 \\
Med-HALT FQT Hallu $\downarrow$ & \textbf{74.81} & 97.67 \\
Med-HALT FCT Accuracy $\uparrow$ & 25.19 & \textbf{33.67} \\
\bottomrule
\end{tabular}
\caption{Reviewer ablation. ReGraM vs.\ \textit{w/o Reviewer} on QA accuracy and hallucination benchmarks. Metrics are computed under our GPT-4o--based automatic evaluation protocol on a 0--100 scale (higher is better for accuracy/similarity; lower is better for hallucination/error rates, $\downarrow$).}
\label{tab:reviewer_summary_main}
\end{table}

% ======================================================4.5 절!!! Efficiency Analysis!!!======================================================
\subsection{Efficiency Analysis}
ReGraM reduces MCQ latency by 46\% via region-first design. SAQ latency slightly increases due to verification. Full results are in Appendix~\ref{app:appendix_C.5}.

% ======================================================4.6절!!! Ablation Studies!!!======================================================
\subsection{Ablation Studies}

To analyze how each structural component contributes to performance, we conducted ablation studies targeting the region selector, multi-hop decomposition, region refinement (MMR), reasoning depth, and their interactions.  

% ======================================================Effect of Domain Prior and Multi-hop Reasoning =============================================
\paragraph{Effect of Domain Prior and Multi-hop Reasoning}
Table~\ref{tab:ablation} compares the impact of removing two core components of the region-first design: the \textit{domain prior} and \textit{multi-hop decomposition}.
Both ablation settings retain the region selector but deactivate either the semantic domain prior used during region construction or the hop-by-hop decomposition used in reasoning.

Under the MCQ setting, removing either component results in relatively mild changes.
In contrast, under the SAQ setting, both ablations lead to substantial performance degradation.
In particular, \textbf{w/o Domain Prior} constructs regions without domain-level semantic guidance, which weakens relation reweighting during region selection and allows semantically irrelevant triplets to dominate the selected region.
As a result, the evidence presented to the reasoner becomes noisy and internally inconsistent, leading to sharp accuracy drops on MedQA (51.03 $\rightarrow$ 2.91) and MMLU-Med (65.45 $\rightarrow$ 5.33).

Similarly, the \textbf{w/o Multihop} setting shows pronounced degradation on multi-step reasoning tasks such as MedQA (51.03 $\rightarrow$ 19.80) and MMLU-Med (65.45 $\rightarrow$ 31.86),
suggesting that single-step reasoning fails to reconstruct compositional dependencies across medical concepts.

Overall, these results indicate that the factual stability and reasoning accuracy of ReGraM are jointly supported by
(1) the \textbf{domain prior}, which stabilizes region construction by suppressing semantically irrelevant evidence,
and (2) the \textbf{multi-hop reasoning} mechanism, which enables structured inference over interdependent medical concepts.

% ======================================================Effect of Reviewer module=========================================================================
\paragraph{Reviewer Ablation: Accuracy vs. Stability Trade-off}
Table~\ref{tab:reviewer_summary_main} summarizes the effect of removing the Reviewer. SAQ Accuracy (avg.) is averaged over four SAQ benchmarks (MedDDx-Expert, MedQA, MMLU-Med, and PubMedQA).
Without the Reviewer, ReGraM experiences catastrophic failure on MedHallu (0\% accuracy, 100\% hallucination), while performance on Med-HALT becomes inconsistent, hallucination increases, but some discriminative accuracy improves (e.g., FCT accuracy).
These results highlight a central insight: the Reviewer enforces conservative generation behavior that reduces hallucination, but may under-answer when KG evidence is limited.
Such discriminative accuracy gains come at the cost of substantially increased hallucination rates, underscoring the trade-off between discriminative confidence and factual reliability.

% ======================================================Interaction Between Domain Prior and Multi-hop Reasoning====================================
\paragraph{Interaction Between Domain Prior and Multi-hop Reasoning}
Removing either the domain prior or multi-hop decomposition significantly reduces accuracy, and removing both leads to catastrophic failure, as shown in Appendix ~\ref{app:appendix_C.5}.

% ======================================================Region Refinement (MMR) and Hop-depth ======================================================
\paragraph{Region Refinement and Hop-depth}
Ablation results on MMR and hop-depth variation are provided in Appendix~\ref{app:appendix_C.2}. The region refinement improves SAQ stability by reducing redundancy, and a hop depth of 3 yields optimal factuality-performance trade-offs.

% ======================================================5장 !!! - Discussion======================================================
\section{Discussion}

ReGraM suggests that instability in medical QA is not only a modeling issue but also a \emph{structural} one.
In particular, how and when the reasoning space is bounded can be as important as the choice of the underlying language model.
Reasoning over a heterogeneous KG without explicit boundaries allows irrelevant relations to intervene,
amplifying semantic drift in multi-hop and generative settings.
By constructing a query-aligned region \emph{before} reasoning,
ReGraM effectively bounds the reasoning space and improves robustness under both MCQ and SAQ settings.

We observe a tension between precision and completeness: strict region bounding may lead to under-answering in low-coverage settings,
whereas dense domains benefit more consistently.
This trade-off is also reflected in adversarial settings (e.g., NOTA),
where strict in-region evidence can bias the model toward a plausible in-region option
unless an explicit abstention mechanism is introduced.
These limitations and broader evaluation caveats are discussed in Section~\ref{sec:limitations}.

% ======================================================6장 !!! - Conclusion======================================================
\section{Conclusion}

This work introduced \textbf{ReGraM}, a region-first KG reasoning framework for improving stability and factuality in medical QA.
ReGraM constructs query-aligned local KG regions and performs reasoning within a bounded region,
mitigating inefficiency and semantic drift in heterogeneous biomedical KGs.
Across seven medical QA benchmarks, ReGraM improves accuracy over KGARevion by 8.04 points in MCQ
and 4.50 points in SAQ, while also reducing hallucinations and improving inference efficiency.
Notably, these gains are achieved using a lightweight 8B-scale language model without task-specific fine-tuning,
highlighting the practical efficiency and deployability of region-first reasoning in resource-constrained medical settings.
ReGraM remains robust under free-form generation and empirically suggests improved scalability
by maintaining a bounded reasoning space during inference, although formal guarantees are left for future work.
By treating the KG as a structural component that shapes reasoning,
ReGraM enables more controllable inference; future work will extend region-first reasoning
to better handle answer-absent queries via explicit abstention strategies.

%====================================================================Limitation====================================================
\clearpage
\section{Limitations}
\label{sec:limitations}

Despite the strong empirical performance of ReGraM, several limitations remain and warrant careful consideration.

\paragraph{Reliance on automatic evaluation protocols}
Our evaluation primarily relies on LLM-based automatic judges rather than clinician-centered assessment.
While this choice enables scalable and consistent benchmarking across diverse datasets,
it cannot fully capture clinical appropriateness, safety, or decision validity in real-world medical contexts.
Moreover, the reported inconsistency score (IS) is computed using a fixed scorer prompt with deterministic decoding,
and its absolute value may vary depending on the evaluator model or prompt design.
Accordingly, IS should be interpreted primarily as a relative comparison under a fixed evaluation protocol.
In future work, we plan to incorporate clinician-in-the-loop evaluation and scenario-based expert review
to more comprehensively assess clinical reliability \citep{10.1093/jamiaopen/ooaf054}.

\paragraph{Dependence on relation semantic representations}
ReGraM relies on PrimeKG as its sole knowledge base and does not introduce new entities or relations.
The effectiveness of region selection therefore depends on how well relation semantics are captured
by the relation description map.
Although we expand and refine this map to better reflect PrimeKG’s relational diversity,
some fine-grained or context-dependent relations may still be underrepresented.
This limitation highlights an open challenge in representing domain semantics
when reasoning is explicitly constrained to a localized region.
Improving automatic or learned relation semantic representations remains an important direction for future work.

\paragraph{Challenges in evidence aggregation at the decision stage}
Even when ReGraM successfully accumulates relevant evidence through region selection and multi-hop reasoning,
final answer selection can occasionally become misaligned with the supporting evidence
(Appendix~\ref{app:qual_compare}, Example~2),
particularly in discrete option selection scenarios.
This observation suggests that robustness in medical QA depends not only on retrieving and verifying high-quality evidence,
but also on how such evidence is aggregated, calibrated, and translated into final decisions.
Improving evidence-aware aggregation and decision calibration remains an important direction
for future research \citep{wen2025know}.

\paragraph{Limited scalability analysis}
ReGraM empirically reduces inference latency by reasoning over compact, query-aligned regions
rather than the full knowledge graph.
However, we do not provide a formal asymptotic analysis with respect to KG size.
Consequently, observed efficiency gains may vary depending on KG structure,
domain distribution, and reasoning depth.
A more rigorous theoretical and system-level analysis of scalability
is left for future work \citep{chen2025gril}.

%==========================================================================ethics statement==================================================
\section{Ethics Statement}

This work aims to improve the factual consistency of medical question answering
systems by constraining reasoning over a biomedical knowledge graph.
All experiments were conducted on publicly available benchmark datasets using
automatic evaluation protocols, and no private or patient-specific data were used.

The generated outputs are not intended for clinical use without human oversight,
and the system should not be deployed in real-world diagnostic settings without
expert review, as medical inaccuracies may still occur.

An anonymized version of the code and additional materials is available at
\url{https://anonymous.4open.science/r/ReGraM-6B41/},
and will be fully released upon acceptance.

\bibliography{custom}

%==================================================================================================
\clearpage
\appendix

\setlength{\abovecaptionskip}{4pt}
\setlength{\belowcaptionskip}{10pt}

\tcbset{
  myprompt/.style={
    enhanced,
    breakable,
    rounded corners,
    colback=white,
    colframe=black!55,
    boxrule=0.4pt,
    fonttitle=\bfseries\small,
    colbacktitle=black!65,
    coltitle=white,
    title style={rounded corners},
    left=10pt, right=10pt, top=9pt, bottom=9pt,
    before skip=12pt,  % 박스 위 간격
    after skip=4pt,    % 박스 아래(캡션 들어가므로 너무 크지 않게)
    parskip=3pt
  }
}
% --- Two-column fixed layout (NO FLOAT => figure numbers won't reorder) ---
\newcommand{\PromptPair}[8]{%
\noindent
\begin{minipage}[t]{0.485\linewidth}
  \begin{tcolorbox}[myprompt, title=#1]
  \small #2
  \end{tcolorbox}
  \captionsetup{type=figure}
  \captionof{figure}{#3}
  \label{#4}
\end{minipage}\hfill
\begin{minipage}[t]{0.485\linewidth}
  \begin{tcolorbox}[myprompt, title=#5]
  \small #6
  \end{tcolorbox}
  \captionsetup{type=figure}
  \captionof{figure}{#7}
  \label{#8}
\end{minipage}
\vspace{12pt}
}
\newcommand{\PromptSingle}[4]{%
\noindent
\begin{tcolorbox}[myprompt, title=#1]
\small #2
\end{tcolorbox}
\captionsetup{type=figure}
\captionof{figure}{#3}
\label{#4}
\vspace{12pt}
}
%=============== Appendix!!!!==============================
\section{Prompt Templates Used in ReGraM}
\label{app:appendix_A}
ReGraM performs prompt-based reasoning across multiple stages, including query decomposition, region selection, reasoning, answer generation, and verification.  
This appendix visualizes the main prompts used at each stage of execution.  
Each prompt is designed according to its functional role, and curly brackets \{\} denote dynamically filled input slots at runtime.

% ===============================================================
\subsection{Reasoning Prompts}
\vspace{-2pt}
\noindent\textit{Prompts for query decomposition, KG interaction, and hop-level synthesis.}
\vspace{6pt}

% ===================== Prompt 1 =====================
\begin{tcolorbox}[myprompt, title=Prompt 1. Query Domain Classification Prompt]
\small
\textbf{System:}  
You are an expert biomedical query classifier.  
Your task is to choose the single most relevant category for the given query from the list below.  
The query may span multiple domains; select the category that best represents the main thrust of the question.

\textbf{Category Options:}
\begin{itemize}[noitemsep, topsep=2pt, leftmargin=*]
  \item \texttt{"GENE\_PROTEIN"}: Questions about genes, proteins, interactions, expression.
  \item \texttt{"DRUG\_THERAPY"}: Questions about drugs, therapies, treatments, side effects, indications.
  \item \texttt{"DISEASE\_SYMPTOM"}: Questions about diseases, symptoms, phenotypes, diagnoses.
  \item \texttt{"PATHWAY\_METABOLISM"}: Questions about pathways, metabolism, or processes.
  \item \texttt{"INTEGRATED"}: Complex questions spanning multiple categories.
\end{itemize}

\textbf{Query:} \{user\_question\}

\textbf{Output (JSON only):}
\begin{quote}\ttfamily\footnotesize
<<JSON\_START>>\\
\{"category": "DRUG\_THERAPY"\}\\
<<JSON\_END>>
\end{quote}
\end{tcolorbox}

\captionsetup{type=figure}
\captionof{figure}{Query Domain Classification Prompt.  
Classifies a question into one of five high-level biomedical domains (GENE\_PROTEIN, DRUG\_THERAPY, etc.).  
The predicted category is later used for relation weighting and KG region selection}
\label{fig:prompt1}

% ===================== Prompt 2 =====================
\begin{tcolorbox}[myprompt, title=Prompt 2. Multi-hop CoT Decomposition Prompt]
\small
\textbf{System:}  
You are a world-class medical reasoning engine.  
Your task is to decompose a complex medical question into up to three logical, answerable sub-questions (hops).  
First, provide your reasoning process (\textbf{Analysis}), then output the sub-questions as JSON.

\textbf{Example:}
\begin{quote}\ttfamily
User Query: "What drug that targets the FLT4 gene also suppresses inflammation?"\\[3pt]
Analysis:\\
1. The primary entity is "FLT4".\\
2. The secondary condition is "inflammation".\\
3. The task is to find a drug connecting them.\\[3pt]
Output (JSON):\\
\{"hops": ["Hop 1: ...", "Hop 2: ...", "Hop 3: ..."]\}
\end{quote}

\textbf{Task:}  
User Query: <<<\{user\_question\}>>>  
Your Response: Provide both “Analysis” and “Final JSON”.
\end{tcolorbox}

\captionsetup{type=figure}
\captionof{figure}{Multi-hop CoT Decomposition Prompt.  
Decomposes a query into up to three logical sub-questions (hops) using a one-shot Chain-of-Thought prompt.  
Serves as the core step for constructing ReGraM’s reasoning path}
\label{fig:prompt2}

% ===================== Prompt 3 =====================
\begin{tcolorbox}[myprompt, title=Prompt 3. Hypothetical Triplet Generation Prompt (HYBRID Mode)]
\small
\textbf{System:}  
You are a cautious biomedical reasoning expert.  
Your goal is to generate logical triplets to answer the given sub-question.

\textbf{Sub-question:} \{hop\_question\}  

\textbf{Known Facts (from KG, may be empty):} \{known\_facts\}

\textbf{Rules:}
\begin{enumerate}[noitemsep, topsep=2pt, leftmargin=*]
  \item Use only the information in “Known Facts”.
  \item If insufficient, generate general, high-level plausible triplets.
  \item Do not invent new entities or overly specific relations.
  \item All generated triplets must use only entities present in $V_q$ and relations from the allowed schema set R.
\end{enumerate}

\textbf{Output (JSON only):}
\begin{quote}\ttfamily\footnotesize
<<JSON\_START>>\\
\{"Triplets": [["entity\_A", "has\_function", "function\_X"], ...]\}\\
<<JSON\_END>>
\end{quote}
\end{tcolorbox}

\captionsetup{type=figure}
\captionof{figure}{Hypothetical Triplet Generation Prompt (HYBRID Mode).  
Used when insufficient KG evidence is retrieved.  
The LLM conservatively generates triplets using only allowed entities and relations while avoiding new entity creation}
\label{fig:prompt3}

% ===================== Prompt 4 =====================
\begin{tcolorbox}[myprompt, title=Prompt 4. Triplet Revision Prompt (Reviewer Loop)]
\small
\textbf{System:}  
You are a fact-checker verifying the factuality and usefulness of the given triplet.

\textbf{Constraints:}
\begin{itemize}[noitemsep, leftmargin=*]
  \item Relation MUST be one of: \{allowed\_relations\}.
  \item Use only entities present in the input.
\end{itemize}

\textbf{Triplet:} \{t\}  

\textbf{Question (context):} \{q\}

\textbf{Output (JSON only):}
\begin{quote}\ttfamily\footnotesize
<<JSON\_START>>\\
\{"Revised\_Triplets": [["Head", "relation", "Tail"]]\}\\
<<JSON\_END>>
\end{quote}
\end{tcolorbox}

\captionsetup{type=figure}
\captionof{figure}{Triplet Revision Prompt (Reviewer Loop).  
Used when a generated triplet is judged as false by the Reviewer.  
Encourages correcting schema-violating relations and removing unnecessary hallucinated content}
\label{fig:prompt4}

% ===================== prompt 5 =====================
\begin{tcolorbox}[myprompt, title=Prompt 5. Hop Answer Synthesis Prompt — KG\_STRICT Mode]
\small
\textbf{System:}  
You are an expert medical reasoner.  
Based \textbf{only} on the “Verified Facts,” provide a direct, concise answer to the sub-question.

\textbf{Sub-question:} \{hop\_question\}

\textbf{Verified Facts:} \{verified\_triplets\}

\textbf{Concise Answer:}
\end{tcolorbox}

\captionsetup{type=figure}
\captionof{figure}{Hop Answer Synthesis Prompt — KG\_STRICT Mode.  
Used when $\geq$ 10 KG facts are available; the LLM generates answers solely from verified evidence without using its internal knowledge}
\label{fig:prompt5}

% ===================== prompt 6 =====================
\begin{tcolorbox}[myprompt, title=Prompt 6. Hop Answer Synthesis Prompt — HYBRID Mode]
\small
\textbf{System:}  
You are a knowledgeable analyst.  
Derive a final answer by reasoning step-by-step using the verified facts.

\textbf{Sub-question:} \{hop\_question\}

\textbf{Verified Facts:} \{verified\_triplets\}

\textbf{Reasoning Steps:}
\begin{quote}\ttfamily
1. Fact: ...\\
2. Fact: ...\\
3. Connection: ...
\end{quote}

\textbf{Final Answer (1 sentence):}
\end{tcolorbox}

\captionsetup{type=figure}
\captionof{figure}{Hop Answer Synthesis Prompt — HYBRID Mode.  
Activated when 1–9 KG facts exist; the LLM performs stepwise reasoning using partial evidence to derive a final answer}
\label{fig:prompt6}

% ===================== prompt 7 =====================
\begin{tcolorbox}[myprompt, title=Prompt 7. Hop Answer Synthesis Prompt — LLM\_GUESS Mode]
\small
\textbf{System:}  
You are an expert biomedical researcher.  
No specific facts were found in the KG. Based on general knowledge, provide a plausible answer beginning with “Based on general knowledge,”.

\textbf{Sub-question:} \{hop\_question\}

\textbf{Plausible Answer:}
\end{tcolorbox}

\captionsetup{type=figure}
\captionof{figure}{Hop Answer Synthesis Prompt — LLM\_GUESS Mode.  
Applied when no relevant KG facts are retrieved; the LLM relies on parametric knowledge and explicitly expresses uncertainty in its answer}
\label{fig:prompt7}

% ===================== prompt 8 =====================
\begin{tcolorbox}[myprompt, title=Prompt 8. Final Answer Synthesis Prompt]
\small
\textbf{System:}  
You are a lead researcher writing a final, conclusive answer for a user.  
Synthesize the reasoning contained in the Evidence Map to answer the Original Question.

\textbf{Original Question:} \{original\_query\}

\textbf{Evidence Map:} \{evidence\_map\}

\textbf{Final Conclusive Answer:}
\end{tcolorbox}

\captionsetup{type=figure}
\captionof{figure}{Final Answer Synthesis Prompt.  
Integrates hop-level reasoning results from the Evidence Map into a coherent final answer, ensuring logical consistency across hops}
\label{fig:prompt8}

% =============================================================================
\subsection{Evaluation Prompts}
\vspace{-2pt}
\noindent\textit{Prompts for automatic evaluation, hallucination detection, and inconsistency scoring.}
\vspace{6pt}

% ===================== prompt 9 =====================
\begin{tcolorbox}[myprompt, title=Prompt 9. SAQ Judge Prompt (LLM-as-a-Judge)]
\small
\textbf{System:}  
You are an expert evaluator grading a submitted answer.  
Compare the “Ground Truth Answer” with the “Model’s Answer.”

\textbf{Task:}
\begin{enumerate}[noitemsep, topsep=2pt, leftmargin=*]
  \item Assess correctness.  
  \item Provide a similarity score (0.0–1.0).  
  \item Output JSON including:
  \begin{itemize}[noitemsep, leftmargin=*]
    \item \texttt{"reasoning"}  
    \item \texttt{"similarity\_score"}  
    \item \texttt{"is\_correct"} (true if score $\geq$ 0.8)
  \end{itemize}
\end{enumerate}

\textbf{Question:} \{query\}  

\textbf{Ground Truth Answer:} \{ground\_truth\_answer\}  

\textbf{Model’s Answer:} \{model\_response\}
\end{tcolorbox}

\captionsetup{type=figure}
\captionof{figure}{SAQ Judge Prompt (LLM-as-a-Judge).  
Evaluates semantic alignment between a model-generated free-form answer and the ground truth using GPT-4o; responses with similarity $\geq$ 0.8 are considered correct}
\label{fig:prompt9}

% ===================== prompt 10 =====================
\begin{tcolorbox}[myprompt, title=Prompt 10. Hallucination Judge Prompt (MedHallu)]
\small
\textbf{System:}  
You are an expert medical fact-checker.  
Determine whether the model’s answer contains hallucination (unsupported claim, contradiction, fabrication) given the knowledge context.

\textbf{Question:} \{query\}  

\textbf{Knowledge Context:} \{reference\_context\}  

\textbf{Model’s Answer:} \{model\_answer\}

\textbf{Output (JSON only):}
\begin{quote}\ttfamily
– supported / unsupported / contradicted
\end{quote}
\end{tcolorbox}

\captionsetup{type=figure}
\captionof{figure}{Hallucination Judge Prompt (MedHallu).  
Used in MedHallu evaluation to determine whether a model response contains hallucination (e.g., unsupported claims or fabricated facts) given reference context}
\label{fig:prompt10}

% ===================== prompt 11 =====================
\begin{tcolorbox}[myprompt, title=Prompt 11. Inconsistency Scorer Prompt (IS)]
\small
\textbf{System:}
You are an evaluator that measures factual inconsistency (hallucination) between two answers to the same medical question.

\textbf{Task:}
Given a \textbf{Question}, a \textbf{Reference Answer} (ground truth), and a \textbf{Model Answer},
output a JSON object with:
\begin{itemize}[noitemsep, leftmargin=*]
  \item \texttt{"is"}: a number in [0,1], where 1 indicates maximally inconsistent / hallucinated and 0 indicates fully consistent.
  \item \texttt{"rationale"}: a brief explanation (1--2 sentences).
\end{itemize}

\textbf{Constraints:}
\begin{itemize}[noitemsep, topsep=2pt, leftmargin=*]
  \item Output \textbf{JSON only}.
  \item Use deterministic evaluation (temperature=0) in our setup.
\end{itemize}

\textbf{Question:} \{query\}

\textbf{Reference Answer:} \{ground\_truth\}

\textbf{Model Answer:} \{model\_answer\}
\end{tcolorbox}

\captionsetup{type=figure}
\captionof{figure}{Inconsistency Scorer Prompt (IS).
Computes a factual inconsistency score in [0,1] between the model answer and the reference answer under SAQ. We use deterministic decoding (temperature=0) and report dataset-level averages (Table~\ref{tab:main_result})}
\label{fig:prompt11}

% ====================================================================Appendix B !!! ====================================================
\section{Implementation Details}
\label{app:appendix_B}
To enhance reproducibility, this appendix details the implementation settings of ReGraM, including model configurations, inference hyperparameters, KG preprocessing pipeline, relation weighting schema, and full KG statistics.

\subsection{Model and Inference Settings}

\subparagraph{Backbone Language Model}  
The core reasoning modules of ReGraM are based on \texttt{Llama-3.1-8B-Instruct} (query decomposition, triplet generation, hop-by-hop reasoning, synthesis).  
We applied 4-bit quantization (NF4, via \texttt{bitsandbytes}) and used \texttt{torch.bfloat16} precision.

\begin{itemize}[leftmargin=*, noitemsep, topsep=0pt]
  \item \textbf{Model:} \texttt{meta-llama/Meta-Llama-3.1-8B-\allowbreak Instruct}
  \item \textbf{Quantization:} 4-bit NF4
  \item \textbf{Precision:} \texttt{torch.bfloat16}
\end{itemize}

\subparagraph{Reviewer Model (Binary Verifier)}  
LoRA-tuned \texttt{Llama-3.1-8B-Instruct}; entity/relation embeddings initialized via RotatE on PrimeKG.

\begin{itemize}[leftmargin=*, noitemsep, topsep=0pt]
  \item \textbf{Base:} \texttt{meta-llama/\allowbreak Meta-Llama-3.1-8B-\allowbreak Instruct}
  \item \textbf{Embedding Init:} RotatE (PrimeKG)
  \item \textbf{LoRA:} rank $r=64$, $\alpha=16$, modules = \texttt{q\_proj}, \texttt{v\_proj}, dropout=0.1
\end{itemize}

\subparagraph{Inference Settings}

\begin{itemize}[leftmargin=*, noitemsep, topsep=0pt]
  \item Temperature = 0.3, Top-$p$ = 0.9
  \item Max tokens: Classification = 256, Reasoning = 512, Synthesis = 1024
\end{itemize}

\textit{Evaluation Phase:}
\begin{itemize}[leftmargin=*, noitemsep, topsep=0pt]
  \item Model: GPT-4o (OpenAI), Temperature = 0.0
\end{itemize}

% ====================Relation Weighting============================
\paragraph{Expanded Relation Description Map}
ReGraM extends the relation description map used in KGARevion
by providing richer natural-language descriptions, domain annotations,
and domain-aware weights for PrimeKG relation types.
This expansion is designed to improve region-aware retrieval accuracy
rather than to introduce new knowledge.
All mappings operate strictly over existing PrimeKG relations.

% =========================================================B.2 KG preprocessing and Retireval========================================
\subsection{Knowledge Graph Preprocessing and Retrieval}
\label{app:kg_retrieval}

This subsection details the implementation of the \textbf{Region Selection} module (Step~(2) in Figure~\ref{fig:architecture}), which constructs a query-aligned KG region for each sub-question.
As shown in Figure~\ref{fig:architecture}, region selection consists of (i) entity linking and expansion, (ii) domain-aware relation weighting, and (iii) weighted MMR re-ranking to select Top-$K$ triplets that form the localized region $G_q$.

% ===============================================================Domain-specific Relation weighting Scheme-------------------------------------
\begin{table*}[!t]
\centering
\small
\renewcommand{\arraystretch}{1.15}
\setlength{\tabcolsep}{5pt}
\begin{tabular}{lccccc}
\toprule
\textbf{Relation} & \textbf{Gene\_Protein} & \textbf{Drug\_Therapy} & \textbf{Disease\_Symptom} & \textbf{Pathway\_Metabolism} & \textbf{Integrated} \\
\midrule
Interacts\_with & 1.5 & 0.8 & 0.6 & 1.0 & 1.2 \\
Targets & 0.8 & 1.5 & 0.8 & 1.0 & 1.3 \\
Treats & 0.6 & 1.5 & 1.2 & 0.8 & 1.1 \\
Causes & 0.5 & 0.7 & 1.5 & 0.8 & 1.0 \\
Expressed\_in & 1.3 & 0.7 & 0.5 & 1.0 & 1.1 \\
Associated\_with & 1.0 & 1.0 & 1.3 & 0.9 & 1.2 \\
Regulates & 1.4 & 0.8 & 0.7 & 1.5 & 1.3 \\
Occurs\_in & 0.9 & 0.8 & 1.0 & 1.2 & 1.1 \\
\bottomrule
\end{tabular}
\caption{\textbf{Relation weighting matrix used in KG region construction}  
Weights $w_r$ are applied during region selection based on the predicted domain (rows: relations, columns: query domains).  
Higher weights prioritize semantically aligned relations during retrieval}
\label{tab:relation_weights}
\end{table*}

%=========================================================================================
\begin{table}[t]
\centering
\small
\begin{tabular}{l @{\hspace{18pt}} r}
\toprule
Dataset & N \\
\midrule
AfrimedQA (MCQ) & 3{,}910 \\
AfrimedQA (SAQ) & 1{,}236 \\
MedQA (USMLE) & 1{,}273 \\
MedDDx & 1{,}769 \\
MedHallu & 1{,}000 \\
PubMedQA & 1{,}000 \\
MedHalt (FCT) & 300 \\
MedHalt (NOTA) & 300 \\
MedHalt (Fake) & 300 \\
\bottomrule
\end{tabular}
\caption{Number of evaluation instances used in our experiments.}
\label{tab:dataset_stats}
\end{table}

%===============================================================
\needspace{6\baselineskip}
\paragraph{Entity Linking and Expansion (Figure~\ref{fig:architecture}-(2), ``Entity Linking'')}

To convert natural language queries into structured, KG-aligned representations, we adopt a hybrid entity linking pipeline that combines symbolic linking with fuzzy matching and synonym expansion.
This step corresponds to the ``Entity Linking'' component in the Region Selection block of Figure~\ref{fig:architecture}.

\begin{itemize}[topsep=4pt, itemsep=2pt, leftmargin=1.3em]
    \item \textbf{Primary linker:} \texttt{scispaCy} using the UMLS Knowledge Base linker (\texttt{en\_core\_sci\_sm}).
    \item \textbf{Fuzzy Matching:} \texttt{rapidfuzz} (Levenshtein ratio $\geq 90$) to recover surface-form variations.
    \item \textbf{Synonym Expansion:} A domain-specific alias dictionary (\texttt{SYNONYM\_MAP}) curated from PrimeKG node attributes (e.g., ``Tylenol'' $\rightarrow$ ``Acetaminophen'').
\end{itemize}

\paragraph{Domain-Specific Relation Weighting (Figure~\ref{fig:architecture}-(2), ``Domain-Aware Relation Weighting'')}
ReGraM assigns domain-specific weights $w_r$ to relation types during region construction.
This corresponds to the ``Domain-Aware Relation Weighting'' component in Figure~\ref{fig:architecture}-(2), and it biases region selection toward relations that are semantically relevant to the predicted query domain.
Table~\ref{tab:relation_weights} summarizes the full weighting scheme.

\paragraph{Weighted MMR Re-ranking and Region Formation (Figure~\ref{fig:architecture}-(2), ``Weighted MMR'')}
Given candidate triplets connected to the expanded entity set, we re-rank them to balance semantic relevance and diversity using weighted Maximal Marginal Relevance (MMR).
This step corresponds to the ``Weighted MMR'' component in Figure~\ref{fig:architecture}-(2) and selects Top-$K$ triplets to form the localized region $G_q$.

\begin{itemize}[topsep=4pt, itemsep=2pt, leftmargin=1.3em]
    \item \textbf{Embedding:}
    \texttt{sentence-transformers/\\all-MiniLM-L6-v2} (384-d)
    \item \textbf{MMR diversity factor:} $\lambda = 0.7$
    \item \textbf{Region size:} $K = 15$
\end{itemize}

Overall, the above pipeline operationalizes Step~(2) in Figure~\ref{fig:architecture} by transforming domain-aware, query-linked triplet candidates into a compact, query-aligned region used as the reasoning space in subsequent modules.

%========================================================Dataset Statistics
\subsection{Dataset Statistics}
\label{sec:dataset_stats}

Table~\ref{tab:dataset_stats} reports the number of evaluation instances used for each dataset in our experiments.
For large-scale datasets, we follow prior work and report results on a randomly sampled subset.

%==========================================C.1 Effect of Removing MMR 표 9!!! =================================
\begin{table}[t]
\centering
\small
\renewcommand{\arraystretch}{1.15}
\setlength{\tabcolsep}{4.5pt} % 열 간격 조정
\begin{tabular}{llccc}
\toprule
\textbf{Dataset} & \textbf{Type} & \textbf{ReGraM} & \textbf{w/o MMR} & $\Delta$ \\
\midrule
MedDDx-Expert & MCQ & 44.44 & 43.06 & $+$1.38 \\
MedQA         & MCQ & 54.38 & 54.80 & $-$0.42 \\
MMLU-Med      & MCQ & 65.72 & 65.66 & $+$0.06 \\
\midrule
MedDDx-Expert & SAQ & 20.81 & 27.60 & $-$6.79 \\
MedQA         & SAQ & 51.03 & 25.30 & $+$25.73 \\
MMLU-Med      & SAQ & 65.45 & 42.30 & $+$23.15 \\
\bottomrule
\end{tabular}
\caption{\textbf{Effect of removing MMR on MCQ and SAQ accuracy}  
MMR reduces redundancy in KG retrieval, especially in SAQ,  
where excessive overlap between triplets amplifies reasoning drift}
\label{tab:mmr_removal}
\end{table}

% ===================================================Appendix C !!!=======================================

\section{Additional Experiments}
\label{app:appendix_C}
This appendix presents additional ablation results and qualitative reasoning examples that complement the main paper.  
These findings further illustrate ReGraM’s reasoning stability, structural effects, and observed failure modes, providing insights for future analyses.

% ====================== Section C.1 ======================
\subsection{Effect of Removing MMR (Extended Results)}
\label{app:appendix_C.1}

ReGraM employs weighted MMR-based triplet re-ranking to reduce redundancy within the KG region and to construct query-centered, refined subgraphs.  
Table~\ref{tab:mmr_removal} reports extended results showing the performance change when MMR is removed.  
In the MCQ setting, MMR removal has minimal impact, while in SAQ, redundant KG candidates cause increased reasoning drift and reduced consistency.

\begin{table}[t]
\centering
\small
\setlength{\tabcolsep}{6pt}
\renewcommand{\arraystretch}{1.15}
\begin{tabular}{lrrrrr}
\toprule
 & \textbf{H=1} & \textbf{H=3} & \textbf{H=5} & \textbf{H=7} & \textbf{H=10} \\
\midrule
MCQ & 33.82 & \textbf{44.44} & 42.33 & 41.67 & 39.33 \\
SAQ & 14.84 & \textbf{20.81} & 20.33 & 21.00 & 18.00 \\
\bottomrule
\end{tabular}
\caption{\textbf{Impact of reasoning hop-depth on MCQ and SAQ performance (MedDDx-Expert)}  
The best results are obtained at Hop 3, while performance drops at larger hop depths due to over-expansion and noise accumulation.}
\label{tab:hop_depth}
\end{table}

\FloatBarrier

% ====================== Section C.2 ======================
\subsection{Number of reasoning steps(H) Analysis}
\label{app:appendix_C.2}

Reasoning hop depth is a key factor for evaluating how multi-hop decomposition affects the accuracy and stability of KG reasoning.  
Table~\ref{tab:hop_depth} presents results for MedDDx-Expert, measuring accuracy at hop depths of 1, 3, 5, 7, and 10.  
Performance peaks at Hop 3, while deeper paths cause noise propagation and increased hallucination frequency.

%===========================================C.3 Interaction Between Region Selector and Decomposition===================
\subsection{Interaction Between Region Selector and Multi-hop Decomposition}
\label{app:appendix_C.3}

ReGraM’s region-first architecture achieves maximum stability when semantic region selection and stepwise reasoning alignment operate complementarily.  
To evaluate their interaction, we conduct an ablation on MedQA (SAQ) by selectively disabling each component.

Table~\ref{tab:region_multihop} presents the result.
Removing either module causes a significant drop in accuracy, while removing both leads to a catastrophic performance collapse—confirming their complementary role in region-first reasoning.

\begin{table}[t]
\centering
\small
\setlength{\tabcolsep}{6pt}
\renewcommand{\arraystretch}{1.15}
\begin{tabular}{lccc}
\toprule
\textbf{Categorization} & \textbf{Multihop} & \textbf{ACC} & $\Delta$ \\
\midrule
O & O & 51.03 & $+$0.00 \\
X & O & 2.91 & $-$46.17 \\
O & X & 19.80 & $-$29.28 \\
X & X & 12.17 & $-$36.91 \\
\bottomrule
\end{tabular}
\caption{Ablation of region selector and multi-hop decomposition on MedQA (SAQ). Disabling either module leads to severe degradation, and removing both causes the largest performance drop.}
\label{tab:region_multihop}
\end{table}

% ====================== Section C.4 ======================
\subsection{Effect of Reviewer Module (Extended Results)}
\label{app:appendix_C.4}

%=====================================================================================================================

\captionsetup{type=table}
\begin{center}
\footnotesize
\setlength{\tabcolsep}{2.5pt}
\renewcommand{\arraystretch}{1.03}

\resizebox{\columnwidth}{!}{%
\begin{tabular}{llcc}
\toprule
\textbf{Group} & \textbf{Dataset / Setting} & \textbf{ReGraM} & \textbf{w/o Reviewer} \\
\midrule
\multirow{4}{*}{\textbf{SAQ Acc}}
& MedDDx-Expert & 20.81 & 27.67 \\
& MedQA         & 51.03 & 24.33 \\
& MMLU-Med      & 65.45 & 41.67 \\
& PubMedQA      & 44.65 & 39.00 \\
\midrule
\multirow{2}{*}{\textbf{MedHallu}}
& ACC / Sim       & 43.50 / 54.70     & 0.00 / 0.00 \\
& Hallu / RError  & 0.8900 / 0.7200   & 1.0000 / 1.0000 \\
\midrule
\multirow{3}{*}{\textbf{Med-HALT}}
& FQT (ACC / Hallu)  & 0.0233 / 0.7481 & 0.0233 / 0.9767 \\
& FCT (ACC / Hallu)  & 0.2519 / 0.7481 & 0.3367 / 0.6833 \\
& NOTA (ACC / Hallu) & 0.1967 / 0.8033 & 0.2100 / 0.7867 \\
\bottomrule
\end{tabular}%
}
\end{center}

\captionof{table}{\textbf{Reviewer ablation across QA and hallucination benchmarks}
SAQ Acc. reports exact-match accuracy. MedHallu reports ACC, semantic similarity (Sim; 0--100), hallucination rate (Hallu), and reasoning error rate (RError). Med-HALT reports ACC and hallucination rate (lower is better)}
\label{tab:ablation_reviewer}

%=====================================================================================================================
To assess the contribution of the Reviewer module in hallucination suppression and factual consistency, we conducted ablation experiments where the module was removed from ReGraM. Table~\ref{tab:ablation_reviewer} summarizes performance across QA and hallucination benchmarks.

While MCQ results show mixed changes—with slight accuracy increase in MedQA ($+$2.29) and minor drops elsewhere—the impact in SAQ settings is substantial. Removing the Reviewer causes severe degradation in MedQA ($-$26.70), MMLU-Med ($-$23.78), and PubMedQA ($-$5.65). Conversely, MedDDx-Expert (SAQ) shows a surprising accuracy increase, suggesting that conservative verification can introduce a trade-off between factual precision and answer completeness when KG evidence is sufficiently dense.

To further evaluate the Reviewer’s effectiveness, we ablated it from ReGraM and evaluated the model on MedHallu and Med-HALT hallucination benchmarks. As shown in Table~\ref{tab:ablation_reviewer}, the model without the Reviewer fails catastrophically on MedHallu (ACC=0, hallucination=100\%), underscoring the role of verification as a supporting mechanism for enhancing hallucination robustness—particularly when KG evidence is sparse or ambiguous.  
While not the primary driver of factual accuracy, the verification layer contributes to stabilizing generation under uncertain contexts within the region-first structure.

%==========================================C.5 Efficiency Analysis==========================
\subsection{Efficiency Analysis (Extended Results)}
\label{app:appendix_C.5}

\captionsetup{type=table}

\small
\begin{tabular}{lcc}
\toprule
\textbf{Setting} & \textbf{KGARevion (s/item)} & \textbf{ReGraM (s/item)} \\
\midrule
MCQ & 1.85 & \textbf{1.00} \\
SAQ & 3.26 & \textbf{3.90} \\
\bottomrule
\end{tabular}
\captionof{table}{Average inference time per item under MCQ and SAQ settings}
\label{tab:efficiency}

% =====================================C.6 Interaction Between Domain Prior and Multi-hop Reasoning=========================================
\subsection{Interaction Between Domain Prior and Multi-hop Reasoning}
\label{app:appendix_C.6}

\small
\renewcommand{\arraystretch}{1.1}
\centering
\begin{tabular}{cccc}
\toprule
\textbf{Domain Prior} & \textbf{Multihop} & \textbf{ACC} & \textbf{$\Delta$} \\
\midrule
O & O & 51.03 & 0.00 \\
X & O & 2.91 & -48.12 \\
O & X & 19.80 & -31.23 \\
X & X & 12.17 & -38.86 \\
\bottomrule
\end{tabular}
\captionof{table}{Interaction between domain prior and multi-hop reasoning on MedQA (SAQ). Removing either component significantly reduces performance, and removing both causes the most severe degradation}
\label{tab:interaction}

%==========================================================C.8============================================================
%=====================================================Domain별 Hallucination 발생률 IR 표!!! - domain_ir==================================
\subsection{Domain-wise Hallucination Trends}
\label{app:domain-wise_hallucination}

To further investigate the relationship between KG coverage and hallucination robustness, we report in Table~\ref{tab:domain_coverage} the domain-wise inconsistency scores (IS) for each dataset under the SAQ setting.  
While domains such as \texttt{Drug\_Therapy} and \texttt{Gene\_Protein} (with relatively high KG coverage) generally exhibit lower hallucination rates,  
this pattern does not hold uniformly—e.g., the \texttt{Integrated} domain has the highest KG coverage but a relatively high mean IS.  
This suggests that although KG coverage may contribute to consistency, it is not the sole determining factor; factors such as evidence redundancy, relation granularity, or prompt design may also play important roles.

\begin{table*}[!t]
\centering
\small
\setlength{\tabcolsep}{4pt}
\renewcommand{\arraystretch}{1.08}
\begin{tabular}{lccccc}
\toprule
\textbf{Dataset} & \textbf{Drug\_Therapy} & \textbf{Gene\_Protein} & \textbf{Disease\_Symptom} & \textbf{Pathway\_Metab} & \textbf{Integrated} \\
\midrule
MedDDx-B & 0.0887 & 0.0455 & 0.0769 & 0.1429 & 0.0784 \\
MedDDx-I & 0.0370 & 0.0679 & 0.0520 & 0.0400 & 0.0000 \\
MedDDx-E & 0.0000 & 0.1198 & 0.0803 & 0.0526 & 0.0347 \\
MedQA     & 0.1014 & 0.0833 & 0.0628 & 0.0000 & 0.2500 \\
AfrimedQA & 0.0806 & --     & 0.0899 & --     & 0.0294 \\
MMLU      & 0.0167 & 0.0417 & 0.0883 & 0.0750 & 0.0800 \\
PubMedQA  & 0.0310 & 0.0000 & 0.0509 & 0.0147 & 0.0054 \\
\midrule
\textbf{KG Coverage (\%)} & 68.14 & 60.30 & 15.66 & 26.31 & 83.52 \\
\bottomrule
\end{tabular}
\caption{Domain-wise hallucination IS vs KG coverage (SAQ). Higher KG coverage correlates with lower hallucination rates, while sparsely represented domains exhibit greater instability.}
\label{tab:domain_coverage}
\end{table*}

% ======================================================C.7========================================================
\subsection{Dataset-wise Hallucination Trends}
\label{app:dataset_ir_figure}

\captionsetup{type=figure}
\begin{center}
\includegraphics[width=\columnwidth]{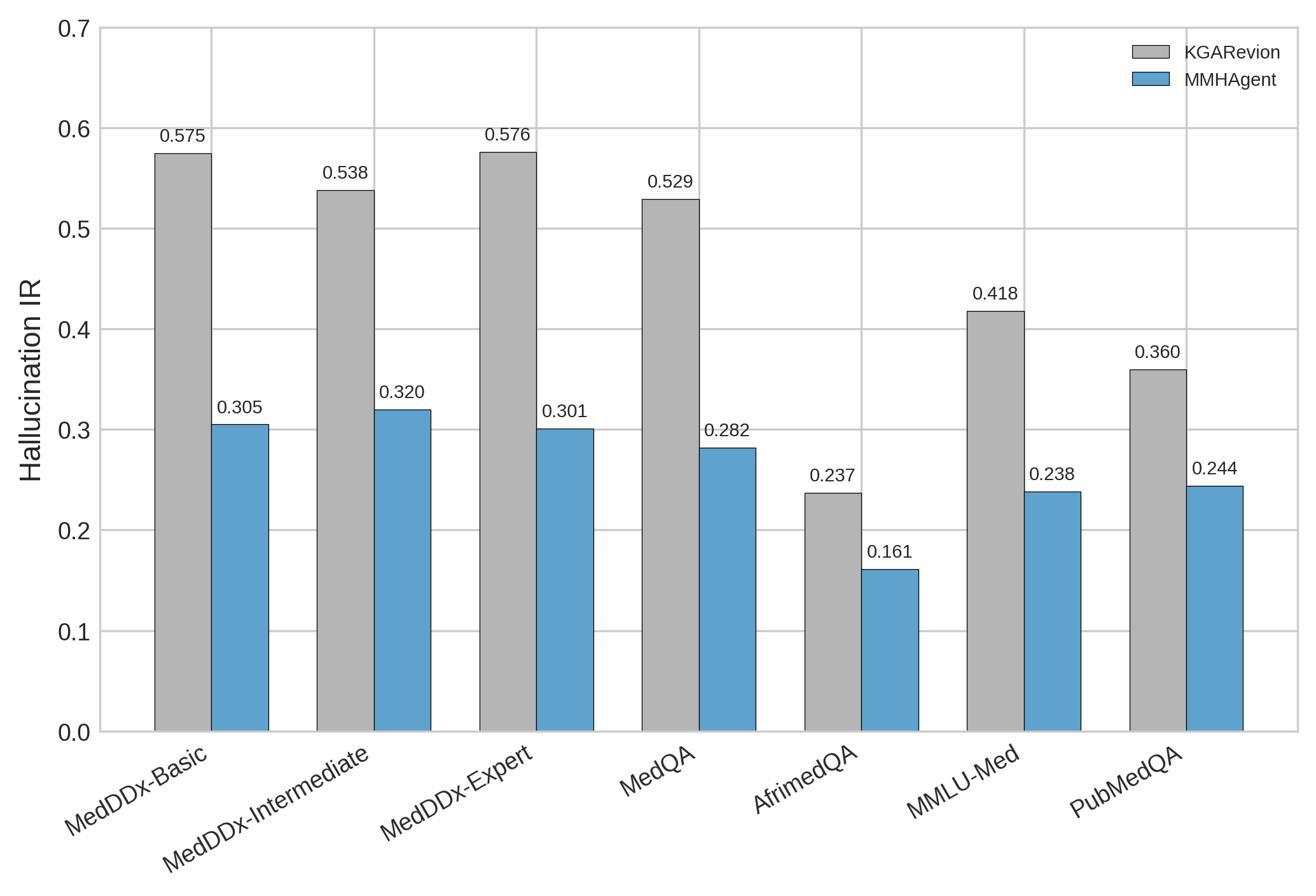}
\end{center}
\captionof{figure}{Reduction in hallucination inconsistency score (IS) across datasets under SAQ setting. ReGraM shows consistent improvement across all benchmarks}
\label{fig:dataset_ir_figure}

%==========================================================C.9============================================================
\subsection{Qualitative Examples (Strengths and a Key Bottleneck)}
\label{app:qual_compare}

We present two step-by-step examples to illustrate (i) when ReGraM succeeds end-to-end in both MCQ and SAQ, and (ii) a representative bottleneck case where multi-hop evidence is correctly accumulated but the final decision step becomes inconsistent with the evidence.

%==================================================== 예시1 : ReGraM 완벽 성공 ==================================================
\noindent\textbf{Example 1 (index=50): End-to-end success in both MCQ and SAQ.}\par\vspace{6pt}
\label{app:example_1}
\begin{tcolorbox}[enhanced, breakable, colback=white, colframe=black!70,
  boxrule=0.4pt, arc=2mm, left=6pt, right=6pt, top=7pt, bottom=6pt]
\small
\textbf{Query.} What are the diseases associated with the NGLY1 gene, and what clinical outcomes does this link entail?\\
\textbf{Options.} (A) ALG1-CDG (B) lipoyl transferase 1 deficiency (C) NGLY1-deficiency (D) aminoacylase 1 deficiency. \textbf{Gold: C}

\vspace{4pt}
\textbf{Step 1: Query typing and decomposition.}
\emph{Domain:} \texttt{GENE\_PROTEIN}. \quad
\emph{Hop decomposition:} single hop (fallback run).

\vspace{4pt}
\textbf{Step 2: Region selection (Top verified triplets).}
\emph{Verified KG triplets (subset):}
(1) (\texttt{NGLY1-deficiency}, \texttt{phenotype present}, \texttt{Developmental regression});\,
(2) (\texttt{NGLY1-deficiency}, \texttt{phenotype present}, \texttt{Cerebral atrophy});\,
(3) (\texttt{NGLY1-deficiency}, \texttt{phenotype present}, \texttt{Alacrima}).

\vspace{4pt}
\textbf{Step 3: Hop-level sub-answer.}
\emph{Sub-answer$_1$:} NGLY1 is associated with \textbf{NGLY1-deficiency}, with neurological/systemic phenotypes such as developmental regression and cerebral atrophy.

\vspace{4pt}
\textbf{Step 4: Final outputs.}
\emph{MCQ:} Pred \textbf{C} (correct). \quad
\emph{SAQ:} Mapped to option \textbf{C} and judged correct under the SAQ protocol.

\vspace{2pt}
\textbf{Takeaway.} ReGraM can solve the same clinical-genetics query end-to-end across both MCQ and SAQ by grounding outcomes in a compact, query-aligned region.
\end{tcolorbox}

%==================================================== 예시2 : 중간 추론은 맞는데 마지막 정합에서 실패 ==================================================
\noindent\textbf{Example 2 (index=49): Evidence is correctly accumulated, but the final decision is inconsistent.}\par\vspace{6pt}
\label{app:example_2}
\begin{tcolorbox}[enhanced, breakable, colback=white, colframe=black!70,
  boxrule=0.4pt, arc=2mm, left=6pt, right=6pt, top=7pt, bottom=6pt]
\small
\textbf{Query.} What are potential diagnoses for chronic cystitis symptoms with reduced bladder capacity and frequent urination?\\
\textbf{Options (MCQ).} (A) cystitis cystica (B) chronic cystitis (C) cystitis (D) interstitial cystitis. \textbf{Gold: D}

\vspace{4pt}
\textbf{Step 0: Hop decomposition (Adaptive\_MultiHop).}
\begin{itemize}[leftmargin=1.2em, itemsep=1pt, topsep=2pt]
  \item Hop 1: candidate diagnoses for chronic cystitis symptoms
  \item Hop 2: diagnoses linked to reduced capacity and frequent urination
  \item Hop 3: final diagnosis matching the combined profile
\end{itemize}

\vspace{3pt}
\textbf{Step 1: Region evidence (verified triplets; subset).}
\begin{itemize}[leftmargin=1.2em, itemsep=1pt, topsep=2pt]
  \item (\texttt{interstitial cystitis}, \texttt{phenotype present}, \texttt{Functional abnormality of the bladder})
  \item (\texttt{interstitial cystitis}, \texttt{phenotype present}, \texttt{Pollakisuria})
  \item (\texttt{interstitial cystitis}, \texttt{phenotype present}, \texttt{Nocturia})
  \item (\texttt{interstitial cystitis}, \texttt{phenotype present}, \texttt{Urinary bladder inflammation})
\end{itemize}

\vspace{3pt}
\textbf{Step 2: Hop-level sub-answers (compressed).}
\begin{itemize}[leftmargin=1.2em, itemsep=1pt, topsep=2pt]
  \item Sub-answer$_1$: Candidate diagnoses include \textbf{interstitial cystitis} and related cystitis subtypes.
  \item Sub-answer$_2$: Evidence links \textbf{interstitial cystitis} to frequent urination and bladder functional abnormality (reduced capacity).
  \item Sub-answer$_3$: The combined symptom profile is most consistent with \textbf{interstitial cystitis} (Option D).
\end{itemize}

\vspace{4pt}
\textbf{Final output (MCQ).} Predicted option: \textbf{B} (incorrect; Gold D).

\vspace{2pt}
\textbf{Takeaway.} Region-first multi-hop reasoning successfully concentrates evidence toward the correct diagnosis (D) at the hop level, but the final discrete decision can still become inconsistent with the accumulated evidence, motivating stronger evidence-aware aggregation and calibrated selection.
\end{tcolorbox}

%=======================================Appendix D - algorithm==============
\section{Region-Constrained Multi-Hop Reasoning Algorithm}
\label{app:algorithm}

% ===================== Algorithm 1 (한 칼럼, inline) =====================
\begin{tcolorbox}[
    title=Algorithm~1: Region-Constrained Multi-Hop Reasoning in ReGraM,
    colback=gray!5,
    colframe=black!80,
    fonttitle=\bfseries,
    boxrule=0.5pt,
    breakable,
    enhanced,
    sharp corners,
    left=6pt,
    right=6pt,
    top=6pt,
    bottom=6pt,
    width=\columnwidth
]
\footnotesize
\textbf{Input.} User query $Q$, knowledge graph $\mathcal{G}$, language model $\mathcal{M}$ \\
\textbf{Output.} Final answer $A$, evidence map $\mathcal{E}$

\vspace{0.6em}
\textit{Phase 1: Query Processing}
\begin{itemize}[leftmargin=1.2em, itemsep=0pt, topsep=0pt]
    \item $D \gets \textsc{DomainClassification}(Q)$
    \item $\{q_1, \dots, q_k\} \gets \textsc{QueryDecomposition}(Q)$ \hfill ($k \le 3$)
    \item $C \gets \emptyset$ \hfill (context: hop-level Q/A memory)
    \item $\mathcal{E} \gets \emptyset$ \hfill (evidence map)
\end{itemize}

\vspace{0.2em}
\textit{Phase 2: Region Construction \& Hop-wise Reasoning (for each sub-question $q_i$)}
\begin{itemize}[leftmargin=1.2em, itemsep=1pt, topsep=1pt]

    % ---- (a) Entity linking / expansion ----
    \item $E_{\text{sci}} \gets \textsc{EntityLink}_{\text{UMLS}}(q_i)$
    \item $E_{\text{fuzzy}} \gets \textsc{FuzzyMatch}(q_i, \mathcal{G})$
    \item $E_{\text{exp}} \gets \textsc{SynonymExpand}(E_{\text{sci}} \cup E_{\text{fuzzy}})$

    % ---- (b) Candidate triplets from the full KG ----
    \item $T_{\text{cand}} \gets \textsc{ExtractTriplets}(E_{\text{exp}}, \mathcal{G})$

    % ---- (c) Region edges (Top-K) and induced subgraph ----
    \item $E_q \gets \textsc{WeightedMMR}(T_{\text{cand}}, D, K=15)$ \hfill (Top-$K$ triplets)
    \item $V_q \gets \{\, v \mid v \text{ appears in a head/tail of some } t \in E_q \,\}$
    \item $G_q \gets (V_q, E_q)$ \hfill (query-aligned reasoning region)

    % ---- (d) Convert region edges to "facts" prompt ----
    \item $F_i \gets \textsc{Textify}(E_q)$ \hfill (facts shown to $\mathcal{M}$)

    % ---- (e) Evidence-aware mode selection (based on region evidence count) ----
    \item $N_{\text{facts}} \gets |E_q|$ \quad and determine mode:
    \begin{itemize}[leftmargin=1em, itemsep=0.5pt, topsep=0.5pt]
        \item If $N_{\text{facts}} \ge 10$, then \textbf{KG-STRICT}
        \item If $0 < N_{\text{facts}} < 10$, then \textbf{HYBRID}
        \item Else, \textbf{LLM-GUESS}
    \end{itemize}

    % ---- (f) Hop reasoning under a hard region constraint ----
    \item $(a_i, T_i) \gets \textsc{Reason}(q_i, F_i, C, \text{mode};\, G_q)$
    \item \textbf{Hard constraint:} all KG lookups / entity expansion inside \textsc{Reason} satisfy \\
    \hspace*{1.2em} $\textsc{Lookup}(\cdot) \subseteq G_q$ and generated entities must be in $V_q$.

    % ---- (g) Verify-and-revise loop only for HYBRID ----
    \item If mode = HYBRID:
    $(a_i, T_i) \gets \textsc{ReviewerVerify}(q_i, a_i, T_i;\, V_q, \mathcal{R})$
    \hfill ($\mathcal{R}$: allowed relations)

    % ---- (h) Update hop memory and global evidence map ----
    \item $C \gets C \cup \{(q_i, a_i)\}$, \quad $\mathcal{E} \gets \mathcal{E} \cup T_i$

\end{itemize}

\vspace{0.2em}
\textit{Phase 3: Final Synthesis}
\begin{itemize}[leftmargin=1.2em, itemsep=1pt, topsep=1pt]
    \item $A \gets \textsc{Synthesize}(Q, C, \mathcal{E})$
    \item \textbf{return} $A, \mathcal{E}$
\end{itemize}
\end{tcolorbox}

\end{document}